\documentclass[11pt,a4paper]{article}
\usepackage[hyperref]{eacl2021}
\usepackage{times}
\usepackage{latexsym}
\usepackage{multirow}
\usepackage{rotating}
\usepackage{verbatim}

\usepackage{microtype}

\aclfinalcopy 

\title{Task-Specific Pre-Training and Cross Lingual Transfer for Code-Switched Data}

\author{Akshat Gupta, Sai Krishna Rallabandi, Alan Black\\
  Carnegie Mellon University \\
  \texttt{akshatgu@andrew.cmu.edu, \{srallaba, awb\}@cs.cmu.edu}}

\date{}

\begin{document}
\maketitle
\begin{abstract}
Using task-specific pre-training and leveraging cross-lingual transfer are two of the most popular ways to handle code-switched data. In this paper, we aim to compare the effects of both for the task of sentiment analysis. We work with two Dravidian Code-Switched languages - Tamil-Engish and Malayalam-English and four different BERT based models. We compare the effects of task-specific pre-training and cross-lingual transfer and find that task-specific pre-training results in superior zero-shot and supervised performance when compared to performance achieved by leveraging cross-lingual transfer from multilingual BERT models. 
\end{abstract}

\section{Introduction}
Code-Switching is a common phenomenon which occurs in many bilingual and multilingual communities around the world. It is characterized by the usage of more than one language in a single utterance \cite{sitaram2019survey}. India is one such place with many communities using different Code-Switched languages, with Hinglish (Code-Switched Hindi-English) being the most popular one. Dravidian languages like Tamil, Kannada, Malayalam and Telugu are also usually code-mixed with English. These code-mixed languages are commonly used to interact with social media, which is why it is essential to be build systems that are able to handle Code-Switched data.

Sentiment analysis poses the task of inferring opinion and emotions of a text query (usually social media comments or Tweets) as a classification problem, where each query is classified to have a positive, negative or neutral sentiment \cite{nanli2012sentiment}. It has various applications like understanding the sentiment of different tweets, facebook and youtube comments, understanding product reviews etc. In times of the pandemic, where the entire world is living online, it has become an even more important tool. Robust systems for sentiment analysis already exist for high resource languages like English \cite{barbieri2020tweeteval}, yet progress needs to be made for lower resource and code-switched languages. 

One of the major bottlenecks in dealing with Code-Switched languages is the lack of availability of annotated datasets in the Code-Mixed languages. To alleviate this problem for Dravidian languages, datasets have been released for Sentiment Analysis in Tamil-English \cite{chakravarthi2020corpus} and Malayalam-English \cite{chakravarthi2020sentiment}. Various shared tasks \cite{patwa2020semeval} \cite{chakravarthi2020overview} have also been accompanied by the release of these datasets to advance research in this domain. 

Models built on top of contextualized word embeddings have recently received huge amount of success and are used in most state of the art models. Systems built on top of BERT \cite{devlin2018bert} and its multilingual variants mBERT and XLM-RoBERTa \cite{conneau2019unsupervised} have been the top ranking systems in both the above competitions. 

In this paper, we train four different BERT based models for sentiment analysis for two different code-switched Dravidian languages. The main contributions of this paper are:
\begin{itemize}
    \item Comparing the effects of Task-specific pretraining and cross-lingual transfer for code-switched sentiment analysis. In our experiments, we find that the performance with task-specific pre-training on English BERT models is consistently superior when compared to multilingual BERT models.
    \item We present baseline results for the Malayalam-English \cite{chakravarthi2020sentiment} and Tamil-English \cite{chakravarthi2020corpus} dataset for a three-class sentiment classification problem, classifying each sentence into positive, negative and neutral sentiments. Our results can be used as baselines for future work. Previous work \cite{chakravarthi2020overview} on these datasets treated the problem as a five-class classification problem. 
\end{itemize}

\section{Related work}
Various datasets for Code-Switched Sentiment Analysis have been released in the last few years, some of which have also been accompanied by shared tasks \cite{patra2018sentiment} \cite{patwa2020semeval} \cite{chakravarthi2020overview} in the respective languages. The shared task released with \cite{chakravarthi2020overview} focused on Sentiment Analysis for Tamil-English and Malayalam-English datasets. The best performing systems for both these tasks were built on top of BERT variants. \cite{chakravarthi2020corpus} \cite{chakravarthi2020sentiment} have provided baseline results for sentiment analysis with a dataset of Youtube comments in Tamil-Englsih and Malayalam-English datasets respectively, using various classification algorithms including Support Vector Machines, Decision Trees, K-Nearest Neighbours, BERT based models etc. In our paper, we use BERT, mBERT \cite{devlin2018bert}, XLM-RoBERTa \cite{conneau2019unsupervised} and a RoBERTa based sentiment classification model \cite{barbieri2020tweeteval} for sentiment classification.

Sentiment classification is usually done by classifying a query into one of three sentiments - positive, negative and neutral. In this work, we perform a three-class classification to be able to leverage the power of the TweetEval sentiment classifier \cite{barbieri2020tweeteval}, which was trained on a dataset of English Tweets. The TweetEval model is a monolingual model trained on an out-of-domain dataset for our task ( the Tamil-English and Malayalam-English datasets are made from scraping Youtube comments). We also use mBERT and XLM-RoBERTa \cite{conneau2019unsupervised} models for classification, which are trained on more than 100 languages and are thus able to leverage the power of cross-lingual transfer for sentiment classification. Previous works \cite{jayanthi2021sj_aj} has also shown that mBERT and XLM-RoBERTa based models achieve state of the art performance when dealing with code-switched Dravidian languages. 

\begin{table}
\centering
\begin{tabular}{cccc}
\hline
 \multicolumn{1}{|p{1.5cm}|}{\centering \textbf{Language}} & \multicolumn{1}{|p{1.2cm}|}{\centering \textbf{Positive}} & \multicolumn{1}{|p{1.5cm}|}{\centering \textbf{Negative}} & \multicolumn{1}{|p{1.2cm}|}{\centering \textbf{Neutral}}\\
\hline
Tam-Eng &  10,559 & 2,037 & 850\\
Mal-Eng &  2,811 & 738 & 1,903\\
Hinglish &  6,616 & 5,892 & 7,492\\
\hline
\end{tabular}
\caption{ \label{Table:Dataset}
Dataset statistics for Tamil-English \cite{chakravarthi2020corpus}, Malayalam-English \cite{chakravarthi2020sentiment} and Hinglish (Sentimix) \cite{patwa2020semeval} dataset. These numbers for the depict the entire dataset which is then divided into train, development and test sets by respective authors.
}
\end{table}

\section{Dataset}
In this paper, we primarily test our models on Tamil-English \cite{chakravarthi2020corpus} and Malayalam English \cite{chakravarthi2020sentiment}. The dataset was collected by scrapping Youtube comments from Tamil and Malayalam movies. All the sentences in the dataset are in the latin script. We also use the Sentimix Hinglish dataset \cite{patwa2020semeval} to leverage cross-lingual transfer from Hinglish. The datasets statistics are summarized in Table \ref{Table:Dataset}. The numbers shown are for the entire dataset which was then split into train, development and test sets by the respective authors. 

\section{Models}
We train sentiment analysis models based on top of four BERT variants:

\begin{itemize}
    \item \textbf{BERT} \cite{devlin2018bert} : The original BERT model was trained using Masked Language Modelling (MLM) and Next Sentence Prediction (NSP) objectives on English Wikipedia and BookCorpus \cite{zhu2015aligning} and has approximately 110M parameters. We use the uncased-base implementation from the Hugging Face library for our work.
    
    \item \textbf{mBERT} \cite{devlin2018bert}: This is a multilingual BERT model trained on 104 languages and has approximately 179M parameters. We again use the uncased-base model for our work.
    
    \item \textbf{XLM-RoBERTa} \cite{conneau2019unsupervised}: This is a multilingual RoBERTa model trained on 100 languages and has approximately 270M parameters. The RoBERTa models were an incremental improvement over BERT models with optimized hyperparameters. The most significant change was the removal of the NSP objective used while training BERT. The XLM-RoBERTa model is trained large multilingual corpus of 2.5TB of webcrawled data. We use the uncased-base XLM-RoBERTa model. 
    
    \item \textbf{TweetEval, a RoBERTa based sentiment classifier}: The paper by \cite{barbieri2020tweeteval} is a benchmark of Tweet classification models trained on top of RoBERTa. We use its sentiment classification model, which is referred to as the TweetEval model in this paper. The sentiment classification model was trained on a dataset of 60M English Tweets. The underlying RoBERTa model was trained on English data and has 160M paramemters. 
\end{itemize}

We use the Hugging Face library implementation of these models. We expect mBERT and XLM-RoBERTa based models to leverage cross-lingual transfer from a large pool of languages it is trained on. The TweetEval model was trained on a dataset of English Tweets for the task of sentiment analysis, but is still out-of domain for Youtube comments datasets. It is important to note that all our chosen models either have a different domain, language or task on which they were initially trained and hence are not directly suitable for the task of code-switched sentiment analysis of Youtube comments.

\subsection{Metrics}
We evaluate our results based on weighted average scores of precision, recall and F1. When calculating the weighted average, the precision, recall and F1 scores are calculated for each class and a weighted average is taken based on the number of samples in each class. This metric is apt as the datasets used are unbalanced. We use the same sklearn implementation \footnote{ \url{https://scikit-learn.org/stable/modules/generated/sklearn.metrics.classification_report.html}} of the weighted average metric as used in \cite{chakravarthi2020corpus} \cite{chakravarthi2020sentiment}. All the numbers shown in the paper are weighted average scores.

\begin{table}
\centering
\begin{tabular}{cccc}
\hline
 \multicolumn{1}{|p{0.8cm}|}{\centering \textbf{Model}} & \multicolumn{1}{|p{1.5cm}|}{\centering \textbf{Precision}} & \multicolumn{1}{|p{1cm}|}{\centering \textbf{Recall}} & \multicolumn{1}{|p{1cm}|}{\centering \textbf{F1}}\\
\hline
Baseline &  0.76 & \textbf{0.80} & \textbf{0.78}\\
BERT & 0.74 & 0.78 & 0.75 \\
mBERT &  0.75 & 0.75 & 0.75 \\
XLM-RoBERTa &  0.76 &  0.76 &  0.76\\
TweetEval & \textbf{0.77} & 0.77  & 0.77  \\
\hline
\end{tabular}
\caption{ \label{Table:MonoMal}
Monolingual Results for Malayalam-English. All scores are weighted average scores. 
}
\end{table}

\begin{table}
\centering
\begin{tabular}{cccc}
\hline
 \multicolumn{1}{|p{0.8cm}|}{\centering \textbf{Model}} & \multicolumn{1}{|p{1.5cm}|}{\centering \textbf{Precision}} & \multicolumn{1}{|p{1cm}|}{\centering \textbf{Recall}} & \multicolumn{1}{|p{1cm}|}{\centering \textbf{F1}}\\
\hline
Baseline &  0.66 & 0.79 & 0.66\\
BERT & 0.74 & 0.74 & 0.74 \\
mBERT &  0.75 & 0.77 & 0.76\\
XLM-RoBERTa & 0.75 & 0.78 & 0.76\\
TweetEval & \textbf{0.76} & \textbf{0.79} & \textbf{0.76}\\
\hline
\end{tabular}
\caption{ \label{Table:MonoTam}
Monolingual Results for Tamil-English. All scores are weighted average scores. 
}
\end{table}

\begin{table*}
\centering
\begin{tabular}{|c||l|l|l||l|l|l||l|l|l|}
  \hline
  \multirow{2}{*}{Train Language} 
      & \multicolumn{3}{c||}{Test Language: Tamil} 
          & \multicolumn{3}{|c||}{Test Language: Malayalam} 
             & \multicolumn{3}{|c|}{Test Language: Hinglish}\\             \cline{2-10}
  & mBERT & xlm-r & TE & mBERT & xlm-r & TE & mBERT & xlm-r & TE \\  \hline
  $English$ & - & - & 0.197 & - & - & 0.345 & - & - & .506\\      \hline
  $Hinglish$ & 0.389 & 0.393 & 0.403 & 0.423 & 0.413 & \textbf{0.473} & - & - & -\\      \hline
  $Tamil$ & - & - & - & 0.389 & 0.523 & 0.427 & 0.379 & 0.321 & 0.432\\      \hline
  $Malayalam$ & 0.562 & 0.504 & \textbf{0.621} & - & - & - & 0.376 & 0.321 & \textbf{0.538}\\      \hline
\end{tabular}
\caption{ \label{Table:Cross}
Zero-shot prediction results for different models trained on the \textit{Train Language}. Here we only report the weighted average F1 scores. We use '-' to represent cells that are Not Applicable to cross-lingual transfer. TE stands for the model built on top of the TweetEval sentiment classification model. xlm-r refers to the XLM-RoBERTa model. 
}
\end{table*}

\section{Experiments}
In this paper, we aim to understand the effects of task-specific pre-training and cross-lingual transfer in improving performance of BERT based models on code-switched datasets. \textit{Task-Specific Pretraining} refers pre-training a model on the same task for which a larger dataset is available and fine-tuning the so-obtained model on the target dataset, which is usually much smaller. In this case, we use the TweetEval model trained for the task of sentiment analysis on a large English corpus, which we fine-tune on the  Dravidian code-mixed datasets. \textit{Cross-Lingual Transfer} is commonly referred to the phenomenon of leveraging features and contextual information learnt from a different set of languages for a task with a new target language. Leveraging cross-lingual transfer is one of the main reasons behind training multilingual BERT models, where we expect multilingual BERT models to perform better on an unknown language when compared to an English BERT model.

\subsection{Monolingual Results}
We first present Monolingual Sentiment Classification Results for Tamil-English and Malayalam-English datasets, as shown in Table \ref{Table:MonoMal} and Table \ref{Table:MonoTam} respectively. The baseline model for Malayalam-Engish is based on mBERT while the baseline model for Tamil-English was a Random Forest classifier as presented in \cite{chakravarthi2020overview}. The Baseline results were trained for a five-way classification problem, hence the weighted average scores have been re-weighted so that they correspond to a three-way classification problem. We also train an English BERT model to act as a baseline for understanding the effects of task-specific pre-training and cross lingual transfer.

We can see that the TweetEval Model improves on the Baseline results. The improvement is quite significant for Tamil-English. The unusually high improvement in the Tamil-English dataset can be due to the fact that the two models are trained for two different problems. The Baseline results are for a five-way classification problem where the F1 scores are re-weighted to only include three classes - positive, negative and neutral. Our models are trained specifically for a three-class classification problem. Our results can provide baselines for future work on sentiment analysis for the Dravidian datasets, which is usually studied as a three-class classification problem. The TweetEval model also performs the best out of all the models tested. We see that the multilingual BERT models perform better than the English BERT models, although the improvement is not very drastic. XLM-RoBERTa consistently performs better than mBERT.

The pre-trained mBERT and XLM-RoBERTa models are trained on multiple languages and in multiple scripts. We expect these models to leverage cross-lingual transfer and perform better than the English BERT model in the code-switched domain. Although we do see improvement in performance by multilingual BERT models over the English BERT models, the improvements are not very drastic. On the other hand, we see a consistently larger improvement due to task-specific pre-training when fine-tuning the TweetEval model on the Dravidian language datasets. We hypothesize two possible reasons for cross-lingual transfer being less effective for the Dravidian datasets. The first is that even though the multilingual BERT models were trained on multiple languages, they were trained in languages in their original scripts. The datasets we are considering contain Malayalam and Tamil in Romanized form. Due to this, the multilingual BERT models do not have representations and contexts of Malayalam and Tamil tokens in Romanized form. Thus the multilingual BERT model has to learn the representation of these new tokens just as an English BERT model would. The second reason is that that although the multilingual BERT models were trained on multiple languages, they were not trained on a code-switched dataset.

\subsection{Cross-Lingual Transfer}
In this section, we look at the zero-shot transfer between the different sets of languages for the above used models for sentiment analysis. The results are shown in Table \ref{Table:Cross}. The first column of Table \ref{Table:Cross} refers to the language the given models were trained on. For Hindi-English (Hinglish), we used the Hinglish Sentimix dataset \cite{patwa2020semeval}. 

We first look at the zero shot transfer of English language for Tamil-English, Malayalam-English and Hinglish datasets (refering to the entire first row of Table \ref{Table:Cross}). This is same as looking at the zero-shot performance of the TweetEval model since it was trained on an English corpus. We see that an English sentiment analysis model has the best zero-shot performance for the Hinglish dataset. We hypothesize that this is because the TweetEval model is an in-domain model for the Hinglish dataset \cite{patwa2020semeval} as the Hinglish dataset is also a Twitter corpus. The Malaylam and Tamil datasets are out of domain since they are made from Youtube comments. 

We can also see that the best zero-shot transfer results are obtained when the TweetEval model is fine-tuned on a linguistically closer language than English for each of the three datasets. For example, the best zero-shot results on the Tamil dataset is achieved when we fine-tune the TweetEval model on Malayalam. Similarly, the best zero-shot results for Tamil are achieved when we fine-tune the TweetEval model on Hinglish. Finally, the best zero-shot results for the Hinglish dataset are achieved when we fine-tune the TweetEval model on Malayalam. These results show that task-specific pre-training is more effective for zero-shot performance and hint at the superiority of task-specific pre-training over cross-lingual transfer. 

In the next experiment, we try to leverage cross-lingual transfer along with task-specific pre-training. To do this, we first fine-tune the TweetEval model on the Hinglish dataset. We expect this fine-tuned model to begin to learn code-switching and recognizing new tokens in Hinglish which are not a part of its vocabulary. To make sure that we don't let the model overfit on the Hinglish dataset, we only fine-tune the model on the Hinglish dataset for 1 epoch. Then we fine-tune this model on the target datasets - Tamil-English and Malayalam-English. The results for this experiment are shown in Table \ref{Table:PreTrainMal} and Table \ref{Table:PreTrainTamil}. Though the combined models show improvements for both datasets, the improvements are not statistically significant. Rigorous experiments to explore this idea will be part of our future work.

\begin{table}
\centering
\begin{tabular}{cccc}
\hline
 \multicolumn{1}{|p{2cm}|}{\centering \textbf{Model}} & \multicolumn{1}{|p{1.5cm}|}{\centering \textbf{Precision}} & \multicolumn{1}{|p{1cm}|}{\centering \textbf{Recall}} & \multicolumn{1}{|p{0.5cm}|}{\centering \textbf{F1}}\\
\hline
XLM-RoBERTa &  0.76 &  0.76 &  0.76\\
TweetEval & 0.77 & 0.77 & 0.77 \\
TweetEval +\\Hinglish & \textbf{0.78} & \textbf{0.78} & \textbf{0.78}\\

\hline
\end{tabular}
\caption{ \label{Table:PreTrainMal}
Comparison between performance of XLM-RoBERTa, TweetEval and TweetEval model pretrained on Hinglish data for the Malayalam-English dataset. 
}
\end{table}

\begin{table}
\centering
\begin{tabular}{cccc}
\hline
 \multicolumn{1}{|p{2cm}|}{\centering \textbf{Model}} & \multicolumn{1}{|p{1.5cm}|}{\centering \textbf{Precision}} & \multicolumn{1}{|p{1cm}|}{\centering \textbf{Recall}} & \multicolumn{1}{|p{0.5cm}|}{\centering \textbf{F1}}\\
\hline
XLM-RoBERTa & 0.75 & 0.78 & 0.76\\
TweetEval & 0.76 & 0.79 & 0.76\\
TweetEval + \\Hinglish & \textbf{0.76} & \textbf{0.79} & \textbf{0.77}\\

\hline
\end{tabular}
\caption{ \label{Table:PreTrainTamil}
Comparison between performance of XLM-RoBERTa, TweetEval and TweetEval model pretrained on Hinglish data for the Tamil-English dataset. 
}
\end{table}

\section{Conclusion}
In this paper we present various experiments to compare the effects of task-specific pre-training and cross-lingual transfer on performance of sentiment classification models for code-switched data. To do so, we check the performance of four BERT models on two code-switched languages - Malaylam-English and Tamil English. We find that task-specific pre-training is superior to cross-lingual transfer for our chosen code-switched datasets. 

The results presented in this paper for four different BERT models can be used as baselines for future work on sentiment analysis for the chosen datasets. We also present first results for the TweetEval \cite{barbieri2020tweeteval} sentiment classification model for code-switched data. 


\bibliography{anthology,eacl2021}
\bibliographystyle{acl_natbib}

\end{document}